\documentclass[11pt, oneside]{article}   	
\usepackage{geometry}                		
\usepackage[margin=0.5in]{caption}
\usepackage[parfill]{parskip}    		
\usepackage{graphicx}				
\usepackage[font=normal,skip=6pt]{caption}
\usepackage{subcaption}
\graphicspath{ {.}}
\usepackage{amssymb}
\usepackage{natbib}
\usepackage{authblk}

\title{\vspace{-2cm}\textbf{Node Specificity in Convolutional Deep Nets Depends on Receptive Field Position and Size}}

\author{Karl Zipser
}
\affil{Redwood Center for Theoretical Neuroscience, U.C. Berkeley\\ California, U.S.A.\\ karlzipser@berkeley.edu}

\begin{document}
\maketitle

%
\begin{abstract}
\noindent
In convolutional deep neural networks, receptive field (RF) size increases with hierarchical depth. When RF size approaches full coverage of the input image, different RF positions result in RFs with different specificity, as portions of the RF fall out of the input space. This leads to a departure from the convolutional concept of positional invariance and opens the possibility for complex forms of context specificity.\end{abstract}
%

%
\section*{Introduction}

The two main components of modern feed forward neural network object classifiers -- convolutional processing and architectural depth -- impose conflicting constraints. Convolutional layers in deep networks are based on the concept that a given node type uses a particular receptive field (RF) to perform the same filtering operation at different parts of the image. But central to deep net architectures is the increase in RF size with increasing hierarchical depth. Inevitably, nodes at higher levels come to have RFs that span a considerable part of the image space for RFs centered in the middle of the image space. When a large convolutional RF is positioned elsewhere, parts of the RF fall off the image space. When this happens, the convolutional assumption breaks down, because in this condition nodes are receiving input for only part of their RFs. Here we explore the consequences of this inherent feature of convolutional deep networks by visualizing individual node selectivity at different RF positions. We find that a given node type can have different RF selectivity at different positions in the image. We propose that convolutional deep neural networks train to take advantage of these consequences of the break-down of convolutional processing in order to develop complex forms of context specificity.

%

%
\section*{Methods}

%
\begin{figure}[!tbp]
    \includegraphics[width=0.8\textwidth]{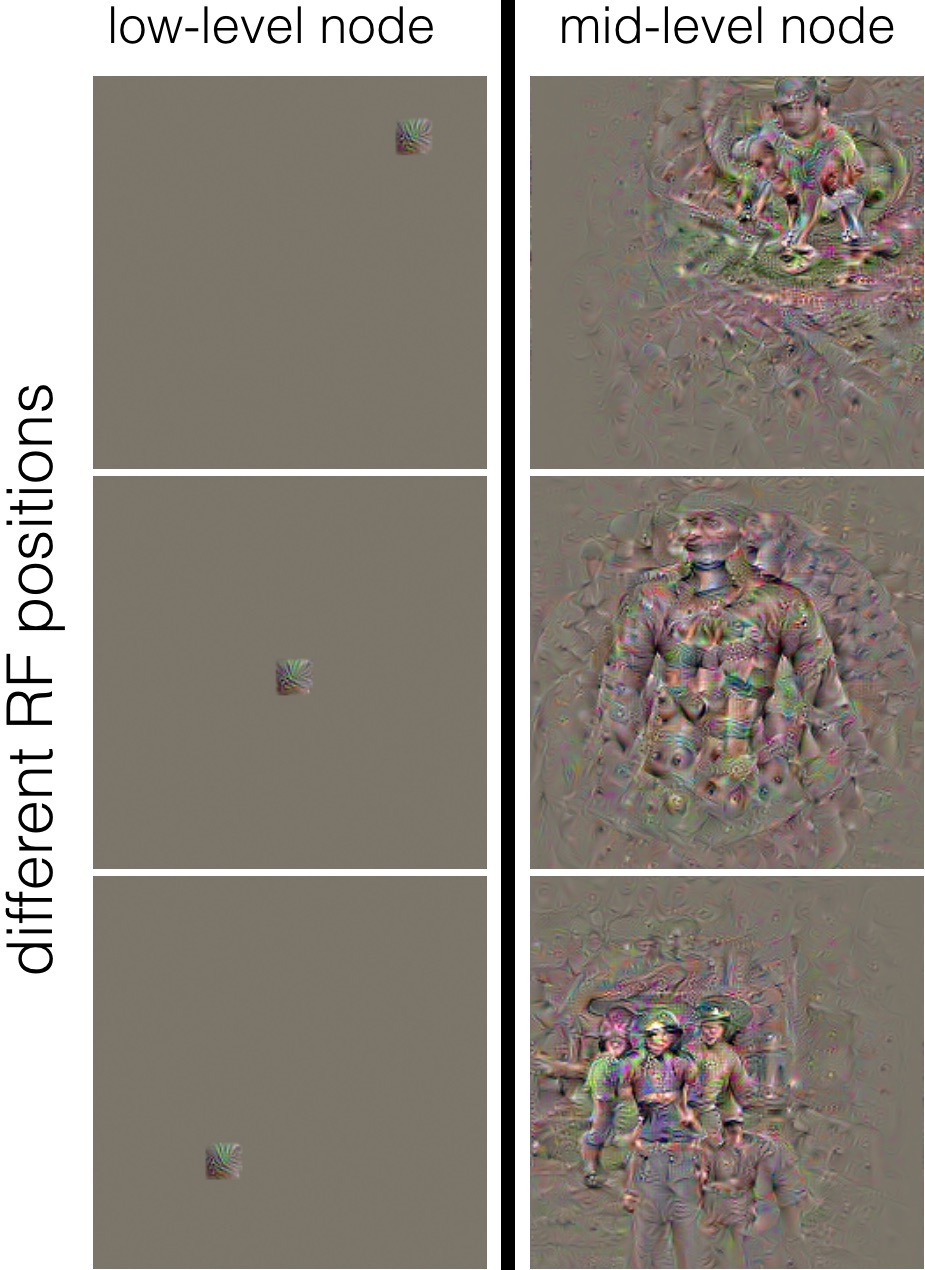}
  \centering
  \caption{How does RF specificity depend on RF position for convolutional nodes? The answer depends on RF size. For nodes with small RFs, there is no position-dependence for the vast majority of RF positions, but for nodes with large RFs, position can have a dramatic effect on RF specificity at most positions.
}
  \label{fig:100}
\end{figure}
%

Visualizing individual node specificity has proven to be an important way to gain insight into deep network functionality \citep{zeiler2014visualizing}. Here we use a method adapted from the Google Inceptionism/Deep Dreams framework  \citep{mordvintsev2015inceptionism}. Gradients for individual convolutional nodes, or all nodes of a given feature class, are set to one, while the gradients for the remaining nodes in the same layer are set to zero. These gradients are backpropagated to the input image. The new input image is the propagated forward, and the process is repeated until stable modified images are generated which are interpreted as the preferred input for the node or nodes in question.

The network studied here is one we trained to categorize 869 person and clothing categories from Imagenet. The network architecture used was GoogLeNet \citep{szegedy2014going} implemented in Caffe as the BVLC GoogleNet (see https://github.com/BVLC/caffe/wiki/Model-Zoo). The network was trained for three million iterations and achieved a top 5 classification success of 67 percent. The relatively high error rate can be attributed to the high degree of ambiguity among the labeled images of people in Imagenet. For example, categories 'old man' and 'grandfather' cannot easily be distinguished.

%

%
\section*{Results}

\subsubsection*{RF specificity as a function of RF position}

How well do nodes maintain their RF specificity across convolutional position in the image? The answer depends on RF size. For nodes with small RFs, visualization of the RF specificity yields instances of the same pattern at different positions in the image -- as expected from the convolution operation. Examples of this are shown in Figure~\ref{fig:100}, left column.

Here the RF of a low level convolutional node type is visualized at three different positions; this visualization reveals an essentially identical pattern for each of these different positions. However, for nodes with large RFs this consistency across RF position does not hold, as seen in the example in Figure~\ref{fig:100}, right column. In the middle row, the RF is centered in the middle of the image, where it shows its highest sensitivity; sensitivity falls off with distance from the RF center, but clearly the RF spans much of the image. When the same node type is visualized for the RF centered in other positions (Figure~\ref{fig:100}, right column, top and bottom), very different RF patterns appear. These patterns are so unlike the pattern for the centered RF that it would be impossible to guess that they all correspond to the same node type.

\subsubsection*{Consistency of visualization outcomes across positions}

\begin{figure}[!tbp]
    \includegraphics[width=1.0\textwidth]{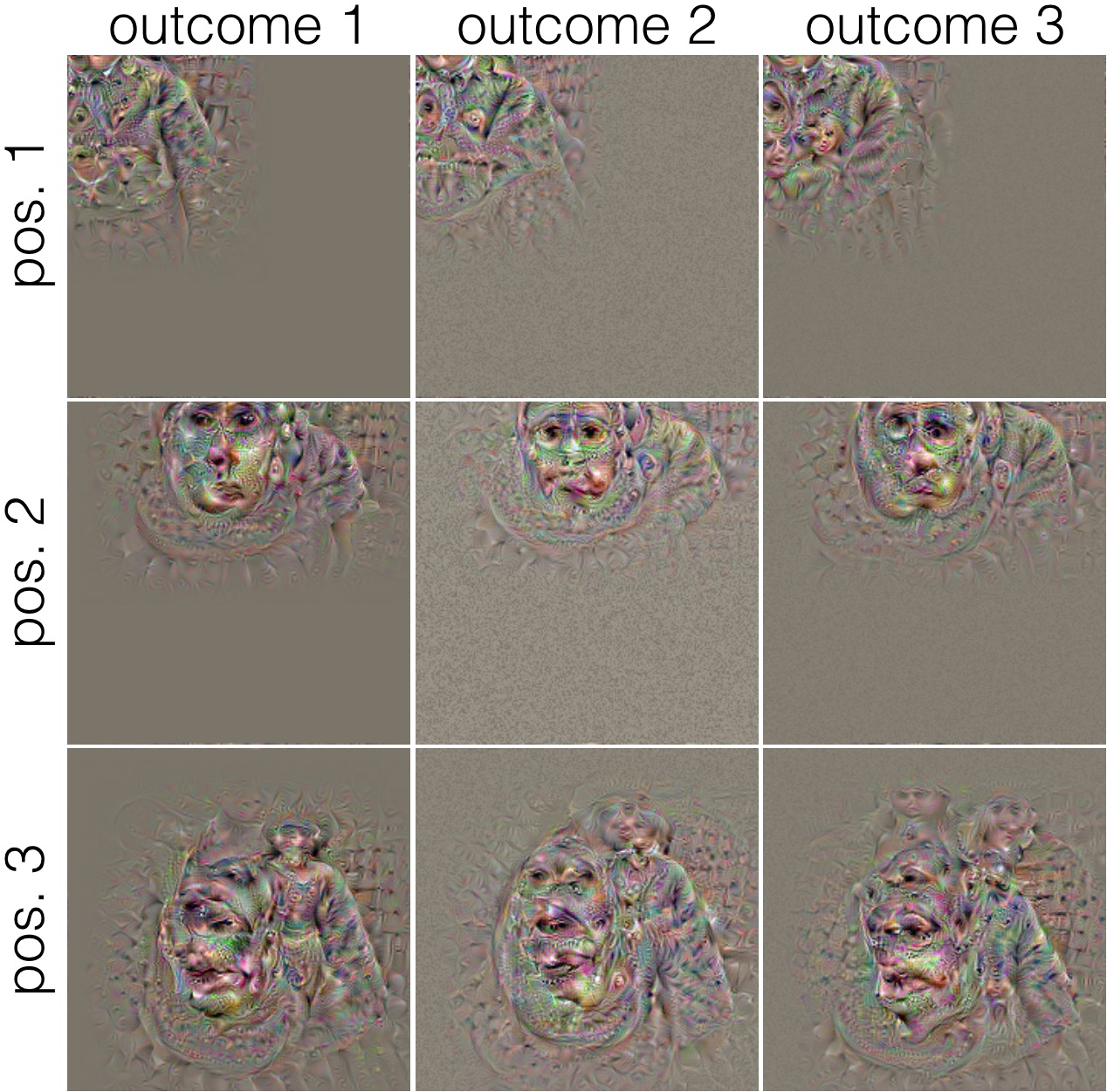}
  \centering
  \caption{Each RF position has a distinct preferred pattern which is largely consistent across visualization outcomes. The different specificities could be interpreted as emphasizing or deemphasizing contextual as opposed to focal elements of the receptive fields.
}
  \label{fig:110}
\end{figure}

Each visualization of an RF is a single outcome of an optimization procedure. An important question is, how different are the separate outcomes for a given RF position, and how does this variability compare to the differences among RF visualizations at different positions? We find that the differential selectivity of convolutional nodes for different RF positions can be quite consistent for a given RF position. For example, in Figure~\ref{fig:110}, RF patterns are examined for a single node, with each row showing three visualization outcomes for a specific RF position.

%
%

Looking across the rows, it is clear that the same fundamental RF pattern appears in the separate outcomes. In the top row, a mid-sized torso appears consistently in each outcome. In the middle row, a large face pointing forward is the consistent outcome. In the bottom row, a large face turned to our left with two small figures in the background, is the consistent outcome. Although each visualization begins with a random dot pattern, and the end results in each case are different on a pixel-by-pixel basis, for this node type the basic semantic and scale structure of the RF patterns are consistent for a given RF position, but are distinct across RF positions.

\subsubsection*{Increasing complexity of RF specificity}

Up to this point visualizations have been presented of nodes at a hierarchical level where the RF patterns are fairly interpretable. With higher level nodes, RF size increases and RF specificity can become much more complex. For example, Figure~\ref{fig:120} shows visualizations of such a node type (comparison to Figures~\ref{fig:100} and~\ref{fig:110} reveals its larger RF extent in the image area). In separate columns, we see extremely complex visualization patterns varying across RF position. In separate rows, we also see much variability across visualization outcomes for a fixed position. Overall, the results indicate both that each RF position yields a range of outcomes, and that the range of these outcomes vary with RF position.

\subsubsection*{Tiling of a node type across position}

\begin{figure}[!tbp]
    \includegraphics[width=1.0\textwidth]{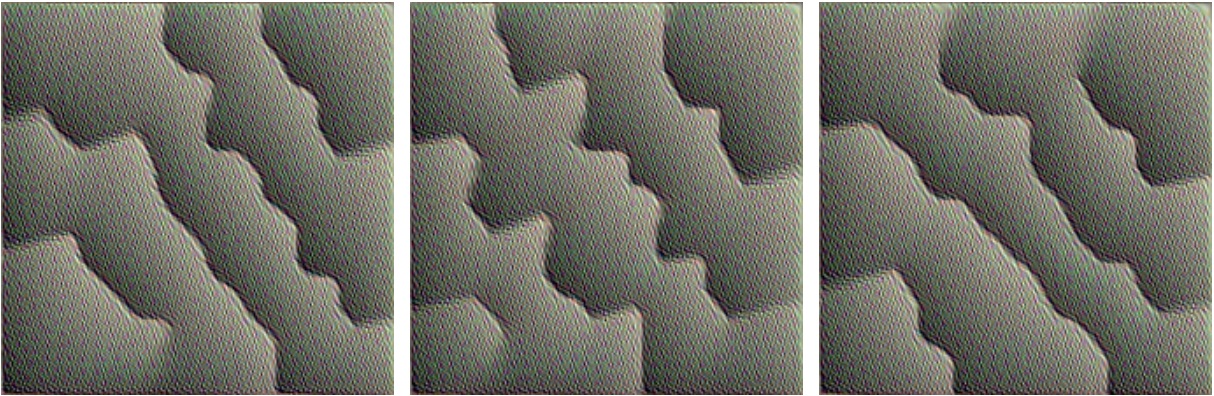}
  \centering
  \caption{Tiled outcomes for low level node with RFs shown in Figure~\ref{fig:100}, left column.
}
  \label{fig:130}
\end{figure}

A complementary approach to visualizing individual RF positions is to generate images that maximize the activation of a given node class at all convolutional positions. Doing this allows us to see the "implications" of a given RF structure which are not intuitively obvious from looking at a single RF pattern, even for a low level node. For example, the low level node in Figure~\ref{fig:100}, left column, when tiled, gives outcomes shown in Figure~\ref{fig:130}. These images display large-scale pattern structure which is highly consistent from outcome to outcome despite variation in local details. Given that small and relatively simple RF structures yield interesting large-scale structure when tiled across all convolutional positions, it becomes interesting to explore what happens when the more complex node types we viewed above are tiled in this manner.

%
\begin{figure}[!tbp]
    \includegraphics[width=1.0\textwidth]{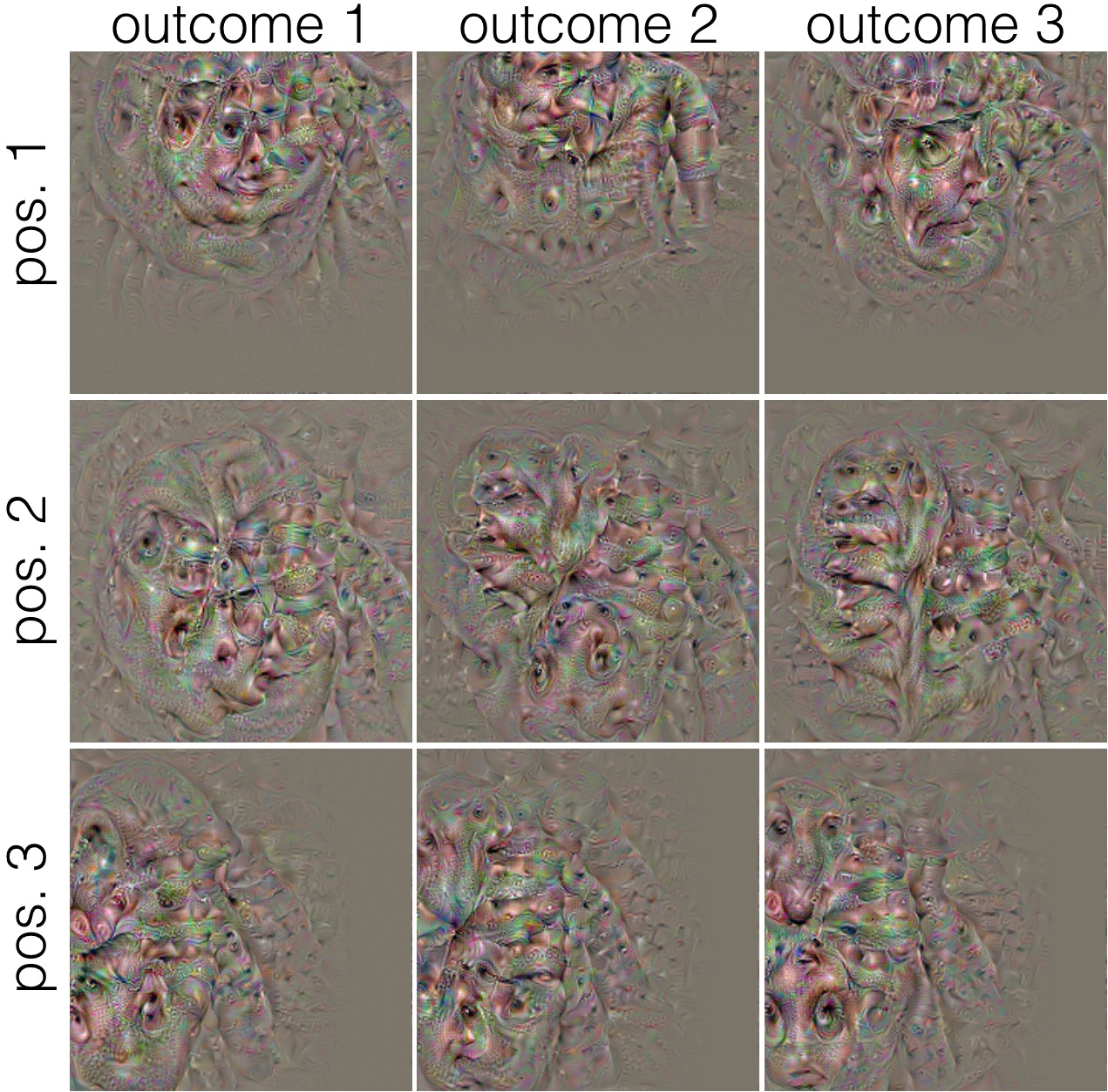}
  \centering
  \caption{Higher level node with more complex preferred patterns, with more variability across visualization outcomes.
}
  \label{fig:120}
\end{figure}
%

We find that tiling more complex convolutional node types typically has three distinct results, two of which are not apparent with low level nodes. First, like low level node types, these nodes generate images with large-scale structure. Second, features are "unmasked" that were not apparent in the individual RF position visualizations. Third, the overall "style" of the resulting images is distinct from the the individual RF position patterns.

%
%

\begin{figure}[!tbp]
    \includegraphics[width=1.0\textwidth]{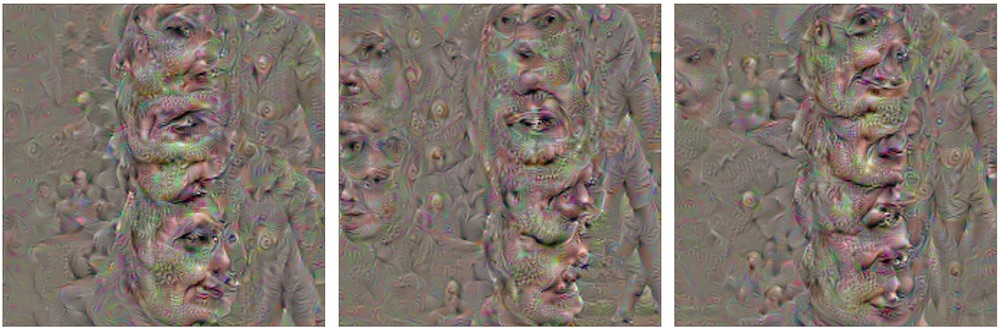}
  \centering
  \caption{Tiled outcomes for mid level node with RFs shown in Figure~\ref{fig:110}.
}
  \label{fig:140}
\end{figure}

All three of these results are evident when we tile the node we saw in Figure~\ref{fig:110}, as shown in Figure~\ref{fig:140}, which reveals an older man in the center, and sometimes produces a somewhat smaller face on the left side; it consistently produces small rows of people in the background to the left, and a medium-sized occluded figure to the right; together, these create large-scale structure in the images. The small rows of people are "unmasked" features not evident in the individual RF position visualizations. The style of these tiled outcomes is quite distinct from that produced by the individual RF visualization; although there are obvious distortions, the recognizable elements are less idealized than the rather naive features in the single RF position visualizations in Figure~\ref{fig:110}. Thus, tiling the node not only generates a larger pattern, but also brings out different features than the individual position visualizations reveals, with a more realistic style. Looking at the single RF visualizations in Figure~\ref{fig:110}, we would not guess that the images in Figure~\ref{fig:140} are the result of tiling them, although some semantic elements and scales are consistent.

\begin{figure}[!tbp]
    \includegraphics[width=1.0\textwidth]{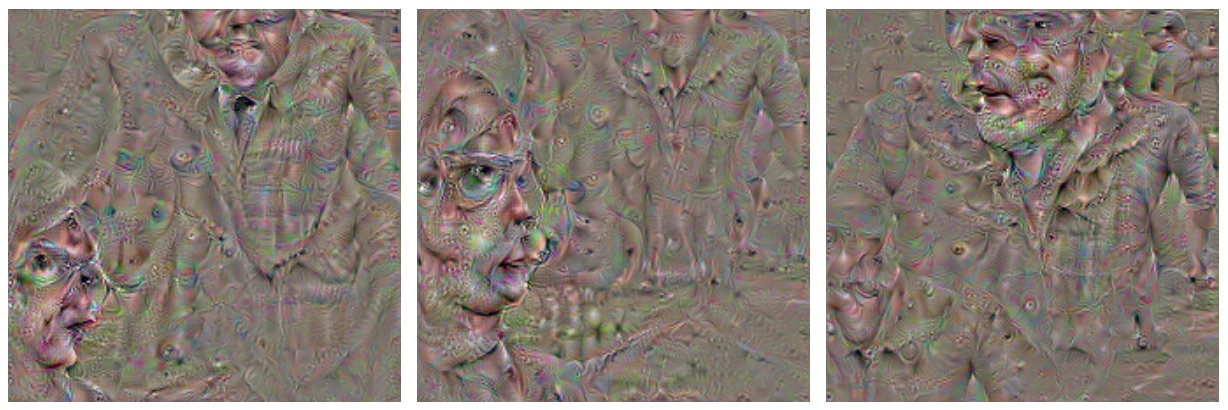}
  \centering
  \caption{Tiled outcomes for mid level node with RFs shown in Figure~\ref{fig:120}.
}
  \label{fig:150}
\end{figure}

The node type we explored in Figure~\ref{fig:120}, when tiled, yields outcomes (shown in Figure~\ref{fig:150}) that are far more interpretable than the individual RF position visualizations of that node type. In place of the extremely busy and confusing patterns in Figure~\ref{fig:120}, we here see largely structured scenes. It is difficult to discern feature unmasking, simply because the individual RF patterns had so many features, but the greater degree of realism is clear in comparison to the fractured images present in Figure~\ref{fig:120}. It is striking that although all individual RF visualizations in Figure~\ref{fig:120} showed a riot of face features, in Figure~\ref{fig:150} the face patterns are quite sparse, generating one or two clear faces per outcome. As a consequence, most of the individual RF positions do not have their preferred feature in the center of their RFs.

%
%

%

%
\section*{Discussion}

When the RF does not fall completely within the image, there can be boundary effects where only part of the RF is stimulated. At earlier convolutional layers with small RFs, this may effect a very small fraction of RF positions. But at intermediate layers, large RFs become the norm. Thus, RFs with parts falling off the image inevitably become the normal functioning condition of the network as a whole. This would seem to indicate that a central idea of convolutional processing -- that the same process is repeated throughout the image -- is not in fact correct for substantial parts of deep convolutional neural networks. In fact, as we have seen above, RF position can have a huge effect on RF selectivity. What are the implications of this? Should this be viewed as failure of the convolutional concept, or is the network able to train and use these properties for enhanced performance? Positional context sensitivity is something the networks could be using this for. Certain types of scenes can have typical large scale structure. Position-modulated RF selectivity may be trained to help detect these types of structures.

%
%

\subsubsection*{Relevance to biological vision and implications for future network design}

The problem of convolutional node RFs "falling off" the image may seem like a specific technical aspect of the current generation of deep networks, but it reflects a general problem in vision -- namely, differential sensitivity of the eye or camera with eccentricity. Biological vision systems have resolution which gradually falls off with eccentricity, instead of an abrupt break at the edge of the image, but the basic problem is the same as with a camera -- there is a point at which sensitivity falls off completely. In the visual system, neurons with large RFs will inevitably have different properties if they are centered on the fovea or on more peripheral eccentricities. Since deep nets are presumably training to deal with edge of image effects, it seems that a more graceful and biological representation of fall off of sensitivity with eccentricity would allow the model to train in a context where convolutional processing could deal with this in a more biologically relevant way. The results could well yield more useful RFs within the models.

\subsubsection*{Relevance to models of visual attention}

The results with tiling have implications for how we conceptualize the role of distributed processing in visual attention. Traditionally, attention is viewed as having the effect of focusing processing in specific locations in spatial or features domains. Our method of visualizing individual RFs can be viewed as an extreme version of this, generating an entire image for the sake of activating a single node type at a single position. However, as we saw in Figure~\ref{fig:120}, this can result in images which are highly confusing and lacking in specific focus. Our method of tiling a given node type over convolutional position would seem to be the opposite of a spatial attention signal -- as it diffuses activation for the node type across the image. And yet this method produces much more focused outcomes, as seen in Figure~\ref{fig:150}. This indicates that the tendency to focus on particular areas of a scene is implicitly trained into the convolutional nodes, and that this tendency is revealed when the nodes are working together -- tiling being an extremely simple form of this. Thus, an attentional mechanism that can direct processing to specific regions or objects of a scene may interact with such a system is a very subtle way, not so much forcing activation to be focused on something specific, but coaxing the system into doing this using the intrinsic properties of the system. This would indicate that models of attention will require much deeper understanding of how convolutional deep networks operate, in order to make full use of their capabilities.

%

\bibliographystyle{natbib}
\bibliography{Zipser}
\end{document}